\documentclass{article}

\usepackage{arxiv}

\usepackage[utf8]{inputenc} 
\usepackage[T1]{fontenc}    
\usepackage{hyperref}       
\usepackage{url}            
\usepackage{booktabs}       
\usepackage{amsfonts}       
\usepackage{nicefrac}       
\usepackage{microtype}      
\usepackage{lipsum}
\usepackage{graphicx}
\graphicspath{ {./images/} }
\usepackage{color,soul,subfigure, subcaption}
\usepackage{multicol}
\usepackage{multirow}

\title{Adinkra Symbol Recognition using Classical Machine Learning and Deep Learning}

\author{\href{https://orcid.org/0000-0003-4426-6137}{\includegraphics[scale=0.06]{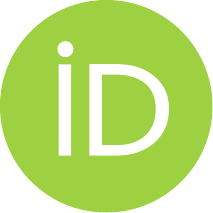}\hspace{1mm}Michael Adjeisah} \\
  National Centre for Computer Animation \\  
  Bournemouth University\\
  Poole, BH12 5BB, United Kingdom \\
  \texttt{madjeisah@bournemouth.ac.uk} \\
  \And
 \href{https://orcid.org/0000-0001-7361-1986}{\includegraphics[scale=0.06]{orcid.pdf}\hspace{1mm}Kwame Omono Asamoah} \\
 Zhejiang Normal University \\
 Jinhua, 321004,  Zhejiang, China \\
\texttt{koasamoah2014@gmail.com} \\
   \And 
 Martha Asamoah Yeboah \\
  College of Computer Science and Technology \\
  Zhejiang Normal University \\
  Jinhua, Zhejiang, 321004 China \\
  \texttt{may138@zjnu.edu.cn} \\
  \And
  \href{https://orcid.org/0000-0003-0909-0826}{\includegraphics[scale=0.06]{orcid.pdf}\hspace{1mm}Raji Rafiu King} \\
  Provincial Key Laboratory of Electronic, Functional Materials and Devices  \\
  Huizhou University \\
  Huizhou City, 516001, Guangdong Province, China \\
  \texttt{mrkingraji@outlook.com} \\
  \And
  \href{https://orcid.org/0009-0007-9472-469X}{\includegraphics[scale=0.06]{orcid.pdf}\hspace{1mm}Godwin Ferguson Achaab} \\
  College of Computer Science and Technology \\
  Zhejiang Normal University \\
  Jinhua, Zhejiang, 321004 China \\
  \texttt{achaabf@gmail.com} \\
  \And
  Kingsley Adjei \\
  College of Computer Science and Technology \\
  Zhejiang Normal University \\
  Jinhua, Zhejiang, 321004 China \\
  \texttt{sleyadjei@gmail.com} \\
}

\begin{document}
\maketitle
\begin{abstract}
Artificial intelligence (AI) has emerged as a transformative influence, engendering paradigm shifts in global societies, spanning academia and industry. However, in light of these rapid advances, addressing the underrepresentation of black communities and African countries in AI is crucial. Boosting enthusiasm for AI can be effectively accomplished by showcasing straightforward applications around tasks like identifying and categorizing traditional symbols, such as Adinkra symbols, or familiar objects within the community.
In this research endeavor, we dived into classical machine learning and harnessed the power of deep learning models to tackle the intricate task of classifying and recognizing Adinkra symbols. The idea led to a newly constructed ADINKRA dataset comprising 174,338 images meticulously organized into 62 distinct classes, each representing a singular and emblematic symbol. We constructed a CNN model for classification and recognition using six convolutional layers, three fully connected (FC) layers, and optional dropout regularization. The model is a simpler and smaller version of VGG, with fewer layers, smaller channel sizes, and a fixed kernel size. Additionally, we tap into the transfer learning capabilities provided by pre-trained models like VGG and ResNet. These models assist us in both classifying images and extracting features that can be used with classical machine learning models. We assess the model's performance by measuring its accuracy and convergence rate and visualizing the areas that significantly influence its predictions. These evaluations serve as a foundational benchmark for future assessments of the ADINKRA dataset. We hope this application exemplar inspires ideas on the various uses of AI in organizing our traditional and modern lives.
\end{abstract}


\section{Introduction}
AI's potential for innovation and advancement is undeniable \cite{stone2022artificial}, with applications ranging from healthcare \cite{shaheen2021applications, secinaro2021role, rajpurkar2022ai, yan2022machine} and transportation \cite{fatemidokht2021efficient, pamucar2022metaverse} to finance \cite{goodell2021artificial,dowling2023chatgpt} and education \cite{zhang2021ai, holmes2021ethics, cope2021artificial}. Nevertheless, in these swift developments, it is imperative to prioritize rectifying the inadequate representation of black communities and African nations within AI. The lack of representation raises concerns about the potential bias and limitations that may arise from predominantly homogeneous perspectives, hindering the realization of AI's full potential and the equitable distribution of its benefits. To comprehensively address this issue, it is necessary to approach it from research areas like Natural Language Processing (NLP) \cite{adjeisah2021pseudotext} and Computer Vision (CV) \cite{islam2022machine, guo2022attention} perspectives, as they entail the utilization of available datasets and establishing baseline models for conducting rigorous experimental analysis.

In the NLP field, most parallel corpora focus on commonly encountered language pairs, such as English-French, -Chinese, -Danish, French-German, and many others. However, there is a noticeable need for parallel corpora for specific language pairs, particularly those involving African languages. This scarcity poses a significant challenge when translating machine translation (MT) \cite{rivera2022machine} from less prominent languages to widely spoken ones. Adjeisah et al. \cite{9443970} elucidated an intricate model for a Bible corpus characterized by extensive parallelism, utilizing Twi, a prevalent Ghanaian language, and extending its application to a select set of languages to fulfill this objective. Such a diverse collection of corpora enables the development of robust MT systems capable of handling multilingual translation tasks effectively \cite{adjeisah2021pseudotext}.

Similarly, examples of computer vision applications are image classification and visual recognition \cite{qiu2021deepsweep, zhang2021deep}. In self-driving cars, deploying neural networks empowers computers to analyze images acquired by onboard cameras. By leveraging this capability, the system can efficiently process visual data, extracting valuable information \cite{lee2023lftk, nixon2019feature} that aids in recognizing and classifying diverse objects and elements encountered on the road. These can include road signs \cite{bayoudh2021transfer, haque2021deepthin}, pedestrians, and vehicles \cite{mo2023pvdet, galvao2023pedestrian}, all contributing to the comprehensive perception and understanding of the surrounding environment of the system. This capability is crucial for autonomous vehicles to make informed decisions and navigate real-time traffic scenarios safely \cite{murthy2022efficientlitedet}. These advanced applications are more commonly found in developed countries, whereas they may appear complex and abstract to individuals in developing countries.

\begin{figure}[!htbp]
     \centering
         \includegraphics[width=10cm]{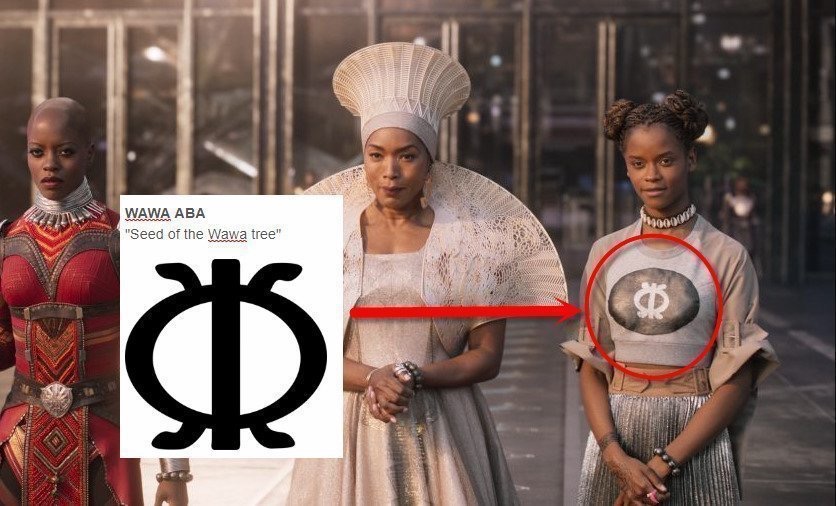}
        \caption{The first costume of the actor with the Wawa motif in the Wakanda movie.}
    \label{fig:wawa}
\end{figure}

Introducing AI to people in developing countries can be presented through simple applications focusing on tasks such as classifying and recognizing traditional symbols, such as Adinkra symbols. Adinkra symbols are integral to Ghanaian culture, representing the people's rich history, values, and beliefs \cite{kuwornu2016philosophy}. These symbols, characterized by visual imagery and proverbs, have a deep cultural significance and are used in various contexts, including textiles, pottery, architecture, and personal adornment \cite{chunfa2021perception}. They serve as emblems of particular thoughts or ideas and are used in many contexts, such as decorative motifs in clothing \cite{blount2022adinkra}, ceramics, and art and design \cite{aidoo2022visual}. The geometric symbols usually comprise a sequence of lines, dots, and other shapes. Every symbol has a distinct meaning, and many of them are connected to certain sayings or proverbs. Common Adinkra symbols include the "Gye Nyame" symbol, which denotes God's supreme might and sovereignty, and the "Sankofa" symbol, which emphasizes learning from the past. The symbols are well known and utilized throughout West Africa and are integral to the cultural legacy of the Ghanaian people. They are an essential component of traditional art and craft and frequently convey concepts and ideas visually \cite{kuwornu2016philosophy, aidoo2022visual}. For example, 
in Black Panther, as depicted in Figure \ref{fig:wawa}, certain symbols were prominently featured, illustrating the vibrant culture of the African community. They were thoughtfully selected to align with the movie's themes. The outfit features an Adinkra symbol. The symbol emerges as the "Wawa" motif\footnote{https://www.insider.com/black-panther-shuri-costume-meaning-2018-5}, which means "the seed of the Wawa tree" and symbolizes toughness, hardness, and perseverance.

Computer vision, a subfield of AI \cite{islam2022machine, guo2022attention}, can significantly classify and recognize these symbols. AI algorithms can be trained on a dataset of adinkra symbols, learning their unique visual features and patterns to classify new symbols into predefined categories or classes. AI can also be utilized for symbol recognition tasks. The system can learn to recognize and match input images or patterns with the corresponding adinkra symbols by training AI models on a large dataset of adinkra symbols. This recognition capability can be applied in various applications, such as translating the adinkra symbol, cultural preservation, or educational purposes. By emphasizing AI's practical and tangible benefits in preserving cultural heritage and facilitating everyday tasks, it becomes more accessible and relatable to individuals in these regions. Demonstrating how AI can be applied in a familiar and meaningful context can help overcome potential barriers and foster greater adoption and understanding of AI technologies.

This study explored the intersection of classical machine learning and advanced deep learning techniques to address the intricate challenge of classifying and recognizing Adinkra symbols within computer vision. Our approach involved creating a novel dataset containing 174,338 images meticulously categorized into 62 distinct classes, each representing a unique and emblematic symbol. The dataset encompasses original images in grayscale and RGB color formats, which effectively represent symbols in diverse visual contexts. However, a noteworthy challenge arises from data imbalance, with certain classes being underrepresented, highlighting the adaptability of machine learning algorithms in the face of such resilience. This imbalance stems from the scarcity of specific symbols, presenting a challenging algorithm scenario.

In our investigation, we used various machine learning models for specific tasks. Support Vector Machine (SVM), K-Nearest Neighbor (kNN), Decision Tree (DT), Random Forest (RF), and gradient-boosted classifiers such as XGBoost were utilized for initial symbol classification and recognition. These algorithms demonstrated their resilience in handling imbalanced data and provided valuable insights into the classification process. 
Additionally, a multi-layer perceptron (MLP) was utilized to capture intricate patterns and connections within the dataset.
For the deep learning aspect, a dual methodology was adopted. Initially, we developed a convolutional neural network (CNN) \cite{alzubaidi2021review} from scratch, leveraging its ability to learn hierarchical features from images automatically. This CNN played a crucial role in extracting intricate features from Adinkra symbols. Furthermore, transfer learning \cite{weiss2016survey} was incorporated by utilizing pre-trained models such as Visual Geometry Group (VGG) \cite{simonyan2014very} and Residual Network (ResNet) \cite{he2016deep}. The pre-trained models underwent fine-tuning on our dataset, leveraging their acquired features to improve classification and recognition accuracy. The integration of classical machine learning and deep learning algorithms demonstrated efficacy in addressing the intricate challenge of Adinkra symbol classification and recognition in varied visual contexts.

Listed below are the significant contributions of this work:
\begin{itemize}
    \item We present a newly constructed ADINKRA dataset comprising 174,338 images with 62 distinct classes, each representing a singular and emblematic symbol.
    \item Implement a CNN model with a more straightforward and smaller version of VGG, with fewer layers, smaller channel sizes, and fixed kernel size layers for classification.
    \item Additionally, we utilize the transfer learning capabilities of pre-trained models such as VGG and ResNet, leveraging them for image classification and feature extraction in classical machine learning models.
    \item We evaluated the performance of the model by measuring its accuracy and convergence rate, along with visualizing the areas that significantly influence its predictions, to serve as a foundational benchmark for future assessment. The source code is available online\footnote{https://github.com/Madjeisah/adinkra-symbols-classification} for reproducibility.
\end{itemize}

The rest of the paper is structured as follows. Section \ref{sec:related} extensively reviews related work in machine learning and deep learning. In Section \ref{sec:method}, we present the core methodology of the investigation. Section \ref{sec:experi} encompasses the description of the data set, the evaluation measurements of the experimental data, and the presentation of the experimental results. Finally, Section \ref{sec:conc} discusses and draws conclusions based on our findings.

\section{Related work} \label{sec:related} 
This part gives an overview of relevant literature about the current study. We briefly examine works within machine and deep learning and explore the literature on transfer learning and general image classification.

\subsection{Machine Learning and Deep Learning}
Machine learning (ML) \cite{mitchell1997machine, jordan2015machine, murphy2022probabilistic} is a formidable tool, empowering computers to acquire knowledge from data and subsequently formulate predictions or decisions \cite{paleyes2022challenges}. As a subset within the domain of artificial intelligence, this field concentrates on formulating algorithms and models that allow computers to learn from data and categorize objects present within the dataset.
ML explicitly involves the use of computational methods to extract patterns \cite{lee2023lftk, nixon2019feature} and insights from the data, allowing machines to improve their performance over a period of time with exposure to more data
\cite{nevatia1980linear, khalid2014survey}. Several types of ML exist, including supervised \cite{burkart2021survey, debnath2021analysis, bohnslav2021deepethogram}, unsupervised \cite{amruthnath2018research, alloghani2020systematic, janiesch2021machine}, semi-supervised \cite{zhu2005semi, zhou2021semi, zhu2022introduction, zheng2022simmatch, song2022graph} and reinforcement learning \cite{kaelbling1996reinforcement, li2017deep, ladosz2022exploration}. 

Deep learning \cite{lecun2015deep, guo2016deep}, a category within machine learning, has transformed various domains by empowering machines to gain insights from data and generate precise predictions through the training of artificial neural networks (ANN) \cite{raji2020novel, otchere2021application} that have multiple layers. In contrast to traditional machine learning, which is based on manually engineered methods for the extraction of patterns and insights from data \cite{nixon2019feature}, deep learning operates through end-to-end learning processes \cite{niu2021review}. Using a deep learning framework, convolutional layers automatically extract pertinent features from images \cite{liu2021efficient}.
Traditional machine learning often involves manually designing algorithms to extract features from images. In deep learning, especially with CNNs, features are automatically learned \cite{hu2021model}. Instead of specifying which features to extract, the network learns the most relevant features during the training process \cite{liu2021efficient}. These features are discovered through the network layers and adapt to the specific task.

Deep learning models, including CNNs, Deep Neural Networks (DNNs) \cite{qiu2021deepsweep, zhang2021deep}, Recurrent Neural Networks (RNNs) \cite{sutskever2011generating, lipton2015critical}, and transformers \cite{vaswani2017attention, wolf2020transformers, khan2022transformers}, consist of numerous layers comprising interconnected nodes known as neurons. Each neuron in these models receives input from the preceding layer, conducts computations, and transmits the output to the subsequent layer. Throughout the training phase, the model iteratively adjusts the weights and biases of the neurons to minimize the disparity between its predictions and the desired output. This optimization process is facilitated by algorithms such as gradient descent \cite{mason1999boosting, bottou2012stochastic, ruder2016overview}. For instance, when presented with an image and a specific task, such as classification, the network autonomously acquires hierarchical representations of the data. This enables the model to discern intricate patterns and features, facilitating accurate predictions.

Like machine learning, deep learning algorithms can be supervised, semi-supervised, or unsupervised. They find application across diverse domains, including but not limited to CV, speech recognition, NLP, healthcare, and recommendation systems \cite{isinkaye2015recommendation}.


\subsection{Transfer Learning}
Transfer learning, an approach within the field of machine learning \cite{weiss2016survey}, wields substantial potency and efficacy by harnessing the knowledge acquired by pre-trained models on extensive datasets and employing it for novel, comparable tasks.
This approach is beneficial in computer vision, where deep learning models such as VGG \cite{simonyan2014very}, ResNet \cite{he2016deep}, and GoogLet \cite{szegedy2015going} have achieved remarkable performance on image classification tasks. 
These models have learned to extract meaningful and discriminative features from images that can be used as a starting point for new image classification tasks. Pre-trained models are predominantly used to fine-tune specific tasks or as feature extractors \cite{chen2021pre}. The former refers to an approach where most of the layers of the pre-trained model remain frozen while enabling the training of only a few top layers for a specific task. This approach allows the model to fine-tune its learned features to better align with a specific dataset. The latter is to use the pre-trained model as a feature extractor by removing the last few layers of the model, which are responsible for the final classification, and serializing the output of the remaining layers as features. These extracted features can be input into traditional machine learning models such as random forests or SVM to perform classification tasks.

The approach offers several advantages, including training a deep neural network from scratch, which can be computationally expensive and require large amounts of labeled data \cite{SRINIVAS2019974}. Models have already learned general features from a large dataset, which can be helpful for tasks with limited training data. Leveraging such capabilities benefits the deep learning community, as these pre-trained models have been extensively studied and optimized with expertise.

\subsection{Image Classification and Recognition}
Image classification is an aspect of computer vision and serves as a foundation on which other tasks such as object detection, semantic segmentation, and image captioning are built \cite{ciregan2012multi, chan2015pcanet}. It involves assigning labels or categories to images based on their content. The ultimate goal in image classification is to develop models that can accurately identify and classify objects or scenes within images \cite{li2013adaptive}. On the other hand, image recognition is the ability of an algorithm to identify and understand objects, places, people, and actions in digital images or videos \cite{traore2018deep}. The significant difference is that image classification uses extracted features to predefine the data into classes or labels. Once objects are predefined with their respective labels, image recognition systems aim to identify specific objects, scenes, or patterns within an image. This could involve recognizing everyday objects, animals, people's faces, text, landmarks, or other visual content. For instance, an image recognition system might determine that an image contains a "tree," "cat," "house," or "car."

Image classification and recognition can be performed using various techniques, including traditional machine learning algorithms, deep learning, and transfer learning models \cite{wang2021comparative}. Traditional machine learning approaches often involve extracting handcrafted features from images and training classifiers on those features \cite{lee2023lftk, nixon2019feature}. It is worth noting that CNNs have demonstrated impressive abilities in tasks like image classification and recognition. What sets them apart is their ability to automatically learn hierarchical representations of the data, as explained in the source \cite{traore2018deep}.
 
\begin{figure}[!htbp]
     \centering
         \includegraphics[width=15cm]{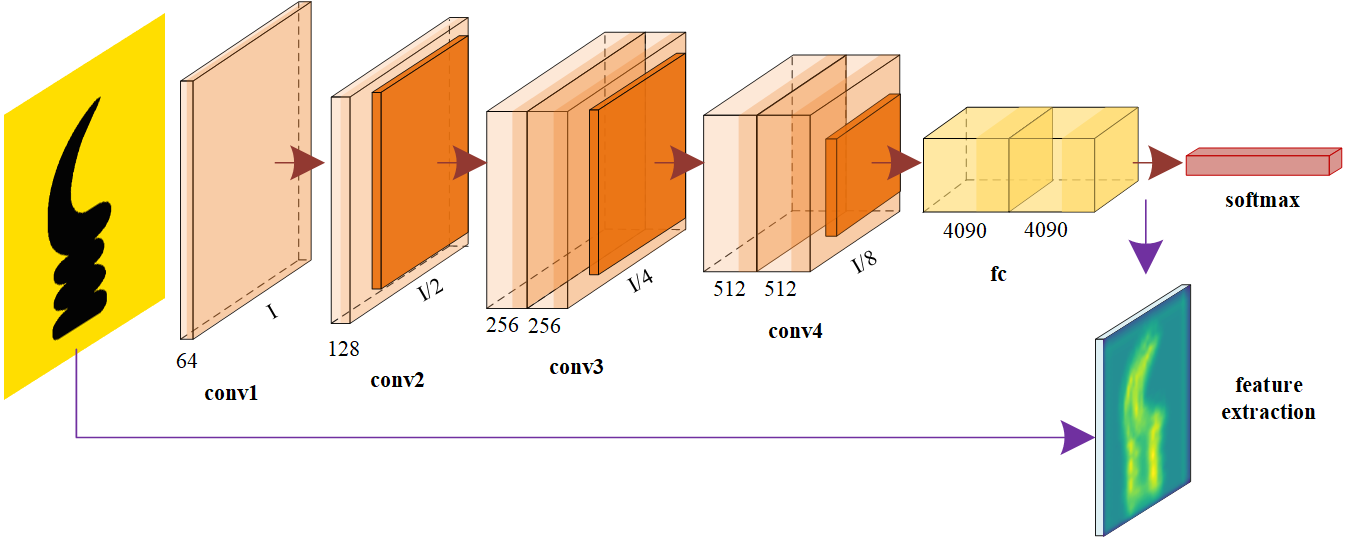}
         \label{fig:y equals x}
        \caption{Custom CNN model architecture as feature extractor and for classification tasks.}
        \label{fig:adin_framework}
\end{figure}

\section{Methodology}\label{sec:method}
This section details the methodology used in the study. It concisely describes the custom model used in this study.

\subsection{Model Description}
This section details the CNN architecture used in this work. Specifically, we describe the custom CNN architecture used for image classification tasks by breaking down each model component. The architecture is inspired by the VGG network \cite{simonyan2014very}, known for its simplicity and effectiveness in image classification tasks. However, our model is a simpler and smaller version than VGG, with fewer layers, smaller channel sizes, and a fixed kernel size of 3x3, as shown in Figure \ref{fig:adin_framework}
A rule of thumb for the convolutional input layer is that the input should be recursively divisible by two. Figure 2 shows the input images with three color channels (typically red, green, and blue, - RGB) and a resolution or input size of 128x128 pixels.
The first convolutional (convo) layer has 64 output channels and a kernel size. We then introduce non-linearity to the model: the Rectified Linear Unit (ReLU) activation function. The second convo layer has 128 output channels and the same kernel, followed by a ReLU activation function. After the first two layers, we introduce a max-pooling layer with a 2x2 window size, reducing the spatial dimensions by taking the maximum value within each 2x2 region. We proceed with a third and fourth convo layer with 256 output channels and a kernel size, each followed by a ReLU, before another (second) max-pooling layer. The fifth and sixth layers have 512 output channels and a kernel size, followed by a ReLU and a final max-pooling layer. We then flatten the output from the sixth layer into a one-dimensional vector. The first two fully connected layers have 4096 units, followed by a ReLU activation, and the final fully connected layer has an output size. There is an optional dropout layer to help prevent overfitting. The model architecture serves as an arbitrary feature extractor and an end-to-end classification task. Given a 2D input image, CNN associates a neuron with the location ($i, j$) to generate a feature value for an output. Mathematically,  this can be computed as:
\begin{equation} \label{eq1}
    \mathcal{A}^l_{i,j,k} = \sigma \Big(\big(\textbf{X}^l_{ij} * \textbf{W}^l_{k} \big) + b^l_{k}\Big) 
\end{equation}
where $\sigma$ denotes an activation function, $\textbf{X}^l_{ij}$ is the input patch centered at surrounding ($i, j$) of the $l$-th layer, $\textbf{W}^l_k$ and $b^l_k$ represent the kernel with learned weight vector and bias term of the $k$-th filter, respectively, and the convolution operation as $*$. The output feature map $\mathcal{A}^l_{:,:,k}$ is used as an input for the next layer by adding a nonlinear activation function to the output of the preceding layer to detect nonlinear features. Let $\eta(\cdots·)$
represent a non-linear function like sigmoid, tanh \cite{lecun2002efficient}, or ReLU \cite{nair2010rectified}, the activation value $\eta^l_{i,j,k}$ of the convo feature map $\mathcal{A}^l_{:,:,k}$ can be expressed mathematically as:
\begin{equation} \label{eq1}
    \eta^l_{i,j,k} = \eta(\mathcal{A}^l_{i,j,k}) 
\end{equation}
The set activation value of the convolutional feature map is relatively large. Therefore, placing a pooling layer between two convo layers is a common practice to attain shift-invariance for dimensionality reduction. The feature maps in a pooling layer are linked to their corresponding feature maps in the prior convolutional layer. This lowers the computational cost of training the network and reduces the risk of overfitting. \textbf{POOL}$(\cdots·)$ is pooling function for various feature maps $\eta^l_{:,:,k}$ in a local neighborhood around the surrounding ($i, j$) in the $k$-th feature map can be expressed as;
\begin{equation} \label{eq1}
    \textbf{\textit{p}}^l_{i,j,k} = \textbf{POOL}(\eta^l_{u,v,k}), \delta(u,v)\in\mathbb{R}_{ij}
\end{equation}
where $\delta(u,v)$ donates the feature product around location ($u,v$) within the pooling region $\mathbb{R}_{ij}$. The output feature value $\textbf{\textit{p}}^l_{i,j,k}$ is then fed to the next layer operation (this could be another convolutional or fully connected (FC) layer). The FC layer is appended towards the end of CNN architecture, with each neuron connected to all neurons of the preceding layer for the classifier. It includes the output of the convolution and pooling phase and a dot product of the weight vector and input vector to obtain the final output.

A network for multi-label problems consists of inputting a set of features originating from the object (e.g., an image). Accurate prediction of the label linked to the object is the main goal. To obtain the optimal network classifier, one must determine the appropriate weights and biases that minimize loss $\mathcal{L}$ between the predicted and the target labels in a training set. A softmax normalization is executed as the final layer to ensure a probability distribution over the output classes. For a training set $\{x^{(i)}, y^{(i)}\}$, such that $i\in1,...,N$ and $y^{(i)}\in1,...,C$, we use the most common categorical cross-entropy loss, which can be expressed mathematically as;
\begin{equation} \label{eq1}
    \mathcal{L} = - \frac{1}{N} \displaystyle \sum^N_{i=1}\displaystyle \sum^C_{j=1} y^{(i)}_j \log \hat{y}^{(i)}_j
\end{equation}
where $x^{(i)}$ represent the $i$-th input patch, $y^{(i)}$ been the target label in the $C$ classes and the predicted label $\hat{y} $ of $j$-th class.

\section{Experiments}\label{sec:experi}
This section comprises the statistics of the Adinkra dataset and a comprehensive set of experiments in two aspects. Firstly, we experiment with traditional machine learning methods with thorough parameter tuning. We provide the accuracy for the learning-type classifiers and their corresponding parameter configurations in the first fold. The second fold is the classification performance of the custom CNN, VGG, and ResNet models with systematic comparison. We perform recognition tasks and show the ground truth label, predicted value, and symbol meaning. We visualize the learned features for model interpretability to understand areas significantly influencing the model’s predictions. We also perform a comparative analysis of model efficiency on the custom CNN and the pre-trained models based on computational resources. The Adinkra dataset is introduced in the following subsection and detailed in the supplementary material.

\subsection{Dataset}
The ADINKRA dataset is a collection of Ghanaian cultural symbols extracted from the Google website. These symbols represent Ghanaian history, values, and beliefs. The dataset includes grayscale and color images, which comprehensively represent the symbols. It contains 139,471 training images and 34,867 test images for machine learning algorithms. The dataset consists of 62 distinct classes, each representing a unique symbol of cultural significance. The labels are in the local dialect (Twi) with English meanings. During the training process, it is expected to split the training dataset into two parts: the training set and the validation set, as shown in Table \ref{train:dataset}. The purpose of this split is to use the validation set to evaluate and validate the performance of the models during training before testing with a separate test set. This allows us to assess the models' learning and make necessary adjustments or improvements. The validation set helps us choose the best model based on its performance and select the optimal set of hyperparameters. Once the models have been trained and validated, they can be tested on the independent test set to assess their final performance.

\begin{table}[!htbp]
\caption{Statistics of the dataset}
\label{train:dataset} 
  \centering
  \label{tb1}
  \begin{tabular}{lcccc}
\hline 
     \multicolumn{5}{c}{\textbf{Adinkra datasets}} \\
    \hline
    \hline
     & Train set & Validation set & Test set & Total\\
     & 104,603 & 34,868  & 34,867 & 174,338\\
     \cmidrule{2-5}
    Class(es) &  \multicolumn{4}{c} {62}   \\
\hline
  \end{tabular}
\end{table}

\subsection{Experimental Settings}
We implemented our CNN framework with PyTorch\footnote{https://pytorch.org/tutorials/beginner/basics/autogradqs\_tutorial.html} and trained it on a GeForce RTX 3090 GPU with 24 GB memory. The model was trained with 50 epochs, a categorical Cross-Entropy loss ($\mathcal{L}$), an Adam optimizer (\textit{Opt}),
a learning rate (\textit{Lr}) 1e-4, training batch size ($\textit{B}_{train}$) of 32, prediction batch size ($\textit{B}_{pred}$) of 4, and 4 number of workers ($\textit{N}_{work}$) as shown in Table \ref{tab:parameter}. The model architecture operates in a dual capacity, functioning as an arbitrary feature extractor and a comprehensive end-to-end solution for tackling classification tasks. As a feature extractor, it adeptly dissects complex data, discerning intricate patterns and relevant attributes from the dataset. We finally engaged different learning-type classifiers for classical machine learning. At the same time, in the context of end-to-end classification, it leverages these learned features to make informed decisions, offering a holistic approach to categorize and distinguish diverse inputs efficiently.

\begin{table}[!htbp]
   \caption{Hyper-parameter settings for the custom CNN and pre-trained models.}
  \label{tab:parameter}
  \centering
  \begin{tabular}{lc}
    \toprule
   \multicolumn{2}{c}{\textbf{Hyper-parameters settings}}  \\
   \midrule
    $\mathcal{L}$    & Cross-Entropy\\ 
    \textit{epoch}   & 50    \\ 
    \textit{opt}     & Adam  \\ 
    \textit{Lr}      &  1e-4 \\ 
    $\textit{B}_{train}$ & 32  \\ 
    $\textit{B}_{pred}$  & 4   \\ 
    $\textit{N}_{work}$  & 4   \\ 
    \bottomrule
  \end{tabular}
\end{table}


\subsection{Classification Results and Analysis}
Table \ref{tab:results} presents classification results that serve as a benchmark for this dataset. Each machine learning and deep learning algorithm was executed five times with shuffled training data to ensure the robustness and reliability of the results. The average accuracy of the test set is reported in these repetitions. This comprehensive approach ensures that the performance metrics reported are representative and less susceptible to random variations.
The Adinkra dataset, a well-established reference dataset, is also included in the benchmark for a side-by-side comparison. This inclusion allows for a direct and insightful assessment of the performance of the algorithms on the dataset in comparison to a widely recognized standard. Such comparisons provide valuable context and aid in measuring the dataset's uniqueness and complexity.
Furthermore, a comprehensive table with detailed explanations of each algorithm's methodology and parameters is available online for those seeking a more in-depth understanding of the algorithms and their underlying mechanisms. This supplementary resource allows researchers and practitioners to dig deeper into the nuances of the algorithms used in the study, facilitating a more comprehensive analysis and interpretation of the benchmark results.

We provide the accuracy percentages for different learning-type classifiers and their corresponding parameter configurations for classical machine learning. Analyzing the accuracy of the classifiers, we can see that the Decision Tree (DT) performs worse on the dataset, with accuracy ranging from 21.06\% to 23.06\%, indicating the need for more complex models or better features. Changing the parameters in kNN has only a marginal impact, and the default configuration performed reasonably well.
XGBoost demonstrates relatively good performance, suggesting the need for careful tuning, and Random Forest shows a wide range of accuracy based on configurations, especially when tuned to an optimal configuration. Some parameter changes led to improved accuracy in MLP and LinearSVC with accuracy of 74.92\% and 71.06\%, respectively. They showed significant improvements with specific configurations, indicating their potential for the Adinkra symbol classification task.

\hspace*{-3em}
\begin{table} 
\caption{Accuracy of different learning-type classifiers and their corresponding parameter configurations}
\label{tab:results} 
  \centering
  \label{tb1}
  \begin{tabular}{lllcc}
\hline 
     Leaning Type & \textrm{\textbf{Classifier}} & \textrm{\textbf{Params}} & \textrm{\textbf{Accuracy {\%}}} \\
    \hline
    \hline
                & \multirow{4}{3em}{kNN}  & default & 47.42 \\
                &                         & n\_neighbors=5  weights=uniform p=1 & 45.55 \\
                &                         & n\_neighbors=10  weights=distance p=2 & 48.28 \\
                &                         & n\_neighbors=5  weights=distance p=2 & 49.42 \\
    \cmidrule{2-4}
                & \multirow{6}{3em}{LinearSVC} & default & 70.05 \\
                &  & loss=squared\_hinge, C=1.0, penalty=l2, max\_iter=500 & 70.77 \\
                &  & loss=squared\_hinge, C=1.0, penalty=l2, max\_iter=100 & 70.20 \\
                &  & loss=hinge, C=1.0, penalty=l2, max\_iter=1000 & 71.06 \\
                &  & loss=hinge, C=1.0, penalty=l2, max\_iter=500 & 70.20 \\
                &  & loss=hinge, C=1.0, penalty=l2, max\_iter=100 & 70.48 \\
    \cmidrule{2-4}
                &  \multirow{9}{3em}{RF} & default & 56.16 \\
                &               & n\_estimators=750, criterion=gini, max\_depth=50 & 62.32 \\
                &               & n\_estimators=1000, criterion=gini, max\_depth=100 & 62.03 \\
                &               & n\_estimators=100, criterion=entropy, max\_depth=None & 52.57 \\
                &               & n\_estimators=750, criterion=entropy, max\_depth=50 & 6146 \\
                &               & n\_estimators=1000, criterion=entropy, max\_depth=100 & 61.60 \\
                &               & n\_estimators=100, criterion=log\_loss, max\_depth=None & 52.57 \\
                &               & n\_estimators=750, criterion=log\_loss, max\_depth=50 & 61.46 \\
    Classical ML&               & n\_estimators=1000, criterion=log\_loss, max\_depth=100 & 61.60 \\
    
    \cmidrule{2-4}
                & \multirow{3}{3em}{DT}     & default & 23.06 \\
                & & criterion=entropy, max\_depth=10, splitter=best & 21.77  \\
                & & criterion=entropy, max\_depth=50, splitter=best & 21.06 \\
    \cmidrule{2-4}
                & \multirow{7}{3em}{XGBoost}   & default & 57.16  \\
                & &  n\_estimators=300, max\_depth=6, objective=multi:softmax & 58.02  \\
                & & n\_estimators=300, max\_depth=12, objective=multi:softmax & 58.30  \\
                & & n\_estimators=100, max\_depth=6, objective=multi:softprob & 57.16  \\
                & & n\_estimators=300, max\_depth=6, objective=multi:softprob & 58.02  \\
                & & n\_estimators=100, max\_depth=12, objective=multi:softprob & 58.22  \\
                & & n\_estimators=300, max\_depth=12, objective=multi:softprob & 58.31 \\
    \cmidrule{2-4}
                & \multirow{8}{3em}{MLP}  & default & 68.48 \\
                &   & hidden\_layer\_sizes=(128, ), activation=relu, max\_iter=200 & 68.76 \\
                &   & hidden\_layer\_sizes=(256, ), activation=relu, max\_iter=200 & 69.91 \\
                &   & hidden\_layer\_sizes=(512, ), activation=relu, max\_iter=200 & 70.63 \\
                &   & hidden\_layer\_sizes=(100, ), activation=tanh, max\_iter=200 & 71.20 \\
                &   & hidden\_layer\_sizes=(128, ), activation=tanh, max\_iter=200 & 69. 34\\
                &   & hidden\_layer\_sizes=(256, ), activation=tanh, max\_iter=200 & 72.49 \\
                &   & hidden\_layer\_sizes=(512, ), activation=tanh, max\_iter=200 & 74.92 \\
    \midrule
    \multirow{1}{6em}{Deep Leaning} & \textrm{CNN} & - & 97.29\\
    \midrule
    \multirow{2}{6.73em}{Transfer Leaning} & \textrm{VGG11} & - & 96.78 \\
    & \textrm{ResNet19} & - & 97.94 \\

\hline
  \end{tabular}
\end{table}

\begin{figure}[!htbp]
     \centering
     \begin{subfigure}
         \centering
         \includegraphics[width=5.9cm]{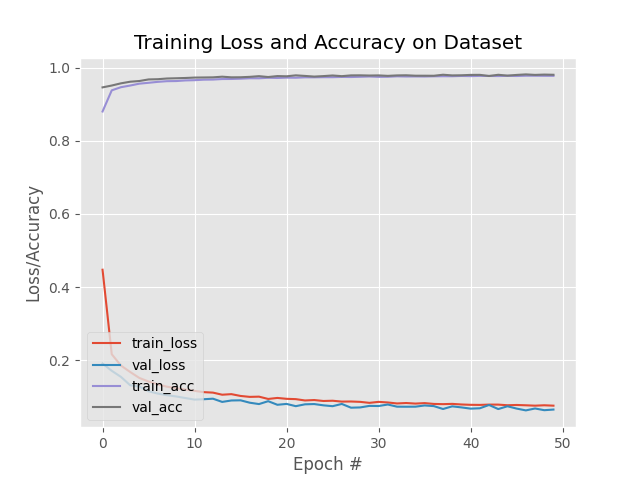}
         \label{fig:three sin x}
     \end{subfigure}
     \hspace*{-2.7em}
     \begin{subfigure}
         \centering
         \includegraphics[width=5.9cm]{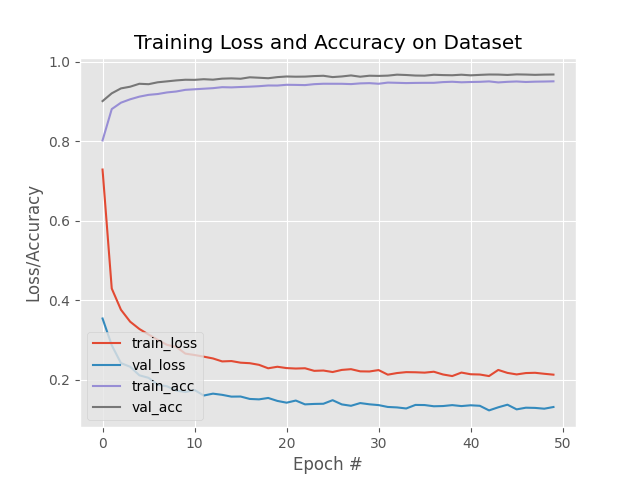}
         \label{fig:five over x}
     \end{subfigure}
     \hspace*{-2.7em}
     \begin{subfigure}
         \centering
         \includegraphics[width=5.9cm]{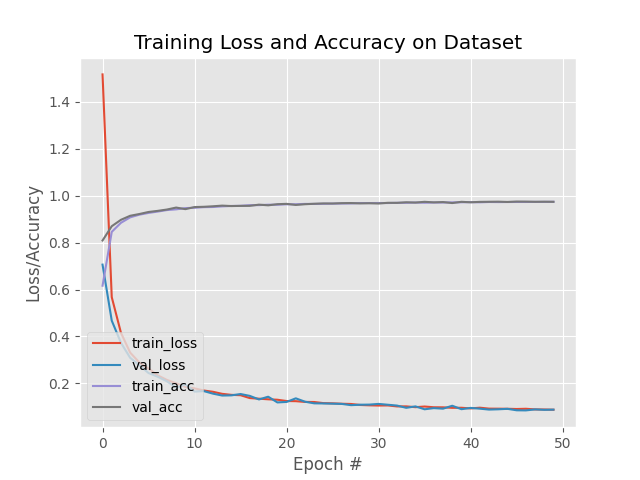}
         \label{fig:five over x}
     \end{subfigure}
        \caption{Model training epochs; Resnet (\textbf{left}), VGG (\textbf{middle}) and our custom CNN (\textbf{right}).}
        \label{fig:model_epoch}
\end{figure}

Our twofold approach for deep learning classification tasks includes a custom CNN and two pre-trained models. The custom CNN model achieved an accuracy of 97.29\%, indicating that it is highly effective. The model architecture designed for the task has learned relevant features directly from the Adinkra data. The model's performance suggests that it understands complex image patterns and features. VGG and ResNet are pre-trained models known for their excellent performance on image classification tasks. ResNet records an accuracy of 97.94\% higher than the custom CNN. This suggests that the architecture and features are well-suited for the classification task, potentially capturing important image details and patterns. Although VGG, on the other hand, is not as high as the custom CNN, it still has a respectable accuracy of 96.78\%, demonstrating that transfer learning from VGG provides valuable features. Ultimately, Figure \ref{fig:model_epoch} indicates that our CNN model has captured the essential patterns in the data and converged well, demonstrating the overall effectiveness of the model's ability to learn from data and generalize its knowledge to make accurate predictions on new, unseen examples during the training process. This phenomenon is associated with the quality of the Adinkra dataset and the quantity of the training data.

In general, all three deep learning models achieved very high accuracy, indicating the effectiveness of the task. However, the difference between pre-trained models and custom CNN depends on factors like available data, computational resources (See Table \ref{train:time}), and the specific requirements of the task. The custom CNN has the advantage of being tailored to the dataset but may require more data and resources for training. On the other hand, pre-trained models offer strong feature extraction capabilities and require less training data and time.

\begin{table}[!htbp] 
\caption{Comparative analysis of model efficiency on the custom CNN and the pre-trained models based on computational resources}
\label{train:time} 
  \centering
  \label{tb1}
  \begin{tabular}{lccc}
\hline 
     \textrm{\textbf{Classifier}} & \textrm{\textbf{Training Time (secs)}} & \textrm{\textbf{Memory (MiB)}} \\
    \hline
    \hline
    \textrm{CNN} & 46227.601 & 16690\\
    \textrm{VGG11} & 15823.055 & 4134\\
    \textrm{ResNet19} & 11638.080 & 1744\\

\hline
  \end{tabular}
\end{table}

\subsection{Adinkra Symbols Recognition}
\begin{figure}[!htbp]
     \centering
     \begin{subfigure}
         \centering
         \includegraphics[width=4cm]{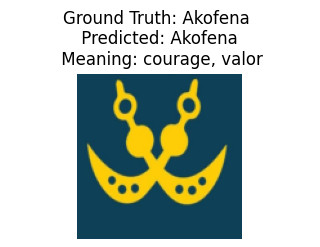}
         \label{fig:y equals x}
     \end{subfigure}
     \hfill
     \hspace*{-3.9em}
     \begin{subfigure}
         \centering
         \includegraphics[width=4cm]{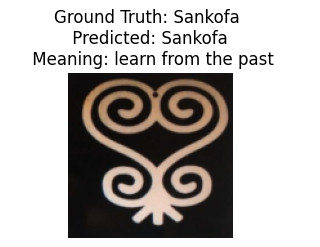}
         \label{fig:three sin x}
     \end{subfigure}
     \hfill
     \hspace*{-3.9em}
     \begin{subfigure}
         \centering
         \includegraphics[width=4cm]{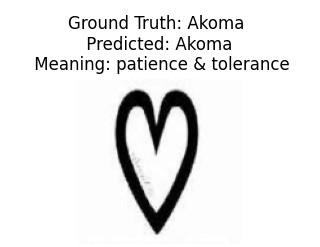}
         \label{fig:five over x}
     \end{subfigure}
     \hfill
     \hspace*{-3.9em}
     \begin{subfigure}
         \centering
         \includegraphics[width=4cm]{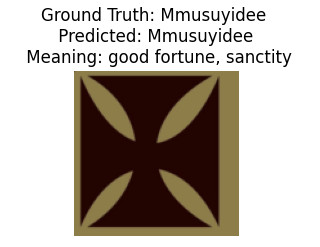}
         \label{fig:five over x}
     \end{subfigure}
     \centering
     \begin{subfigure}
         \centering
         \includegraphics[width=4cm]{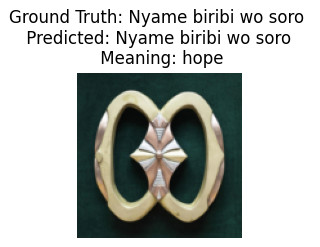}
         \label{fig:y equals x}
     \end{subfigure}
     \hfill
     \hspace*{-3.9em}
     \begin{subfigure}
         \centering
         \includegraphics[width=4cm]{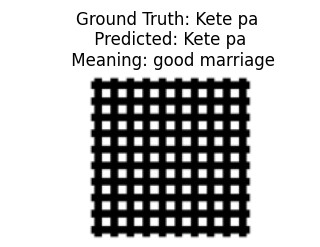}
         \label{fig:three sin x}
     \end{subfigure}
     \hfill
     \hspace*{-3.9em}
     \begin{subfigure}
         \centering
         \includegraphics[width=4cm]{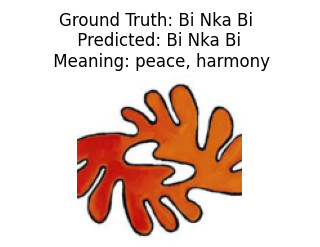}
         \label{fig:five over x}
     \end{subfigure}
     \hfill
     \hspace*{-3.9em}
     \begin{subfigure}
         \centering
         \includegraphics[width=4cm]{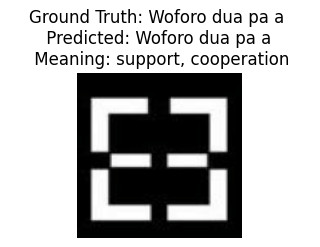}
         \label{fig:five over x}
     \end{subfigure}
        \caption{Recognition task showing the ground truth label, predicted value, and the meaning of the recognized Adinkra symbol.}
        \label{fig:adin_rec}
\end{figure}

Reaching a 97\% accuracy level on a model classifying 62 labels is significant. Nevertheless, it is important to interpret the results with a deeper understanding of what this accuracy represents and its potential implications. Therefore, we successfully perform an object recognition task. Figure \ref{fig:adin_rec} shows various recognition tasks in which we show the ground truth label, predicted value, and meaning of the symbol. This demonstrates that the model's convolutional layers have effectively learned to capture meaningful patterns and extract relevant features from the input images, enabling it to distinguish and recognize the Adinkra symbols. Such success reflects the quality of the training data. The high-quality, diverse, and representative Adinkra training data significantly impacts the model's performance. Therefore, it implies the ability of the model to generalize well by applying its learned knowledge to new, unseen data. 

\subsubsection{Interpretability}
\begin{figure}[!htbp]
     \centering
     \begin{subfigure}
        \centering
        \includegraphics[width=15cm]{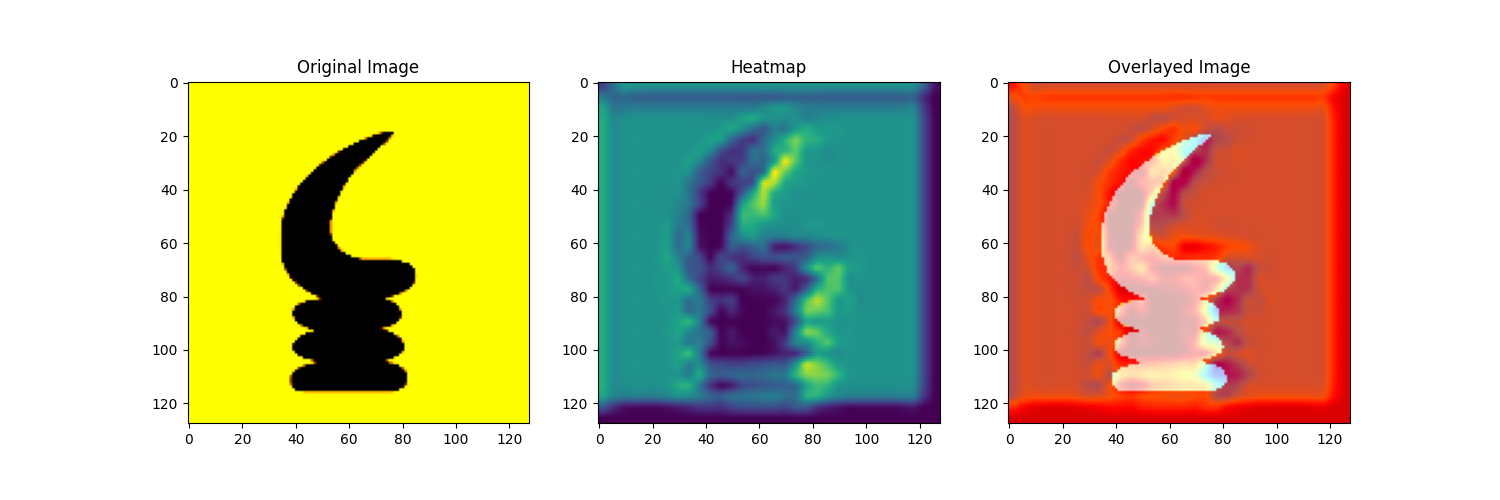}
    \end{subfigure}
    \vspace{-1.7em}
    
     \begin{subfigure}
        \centering
        \includegraphics[width=15cm]{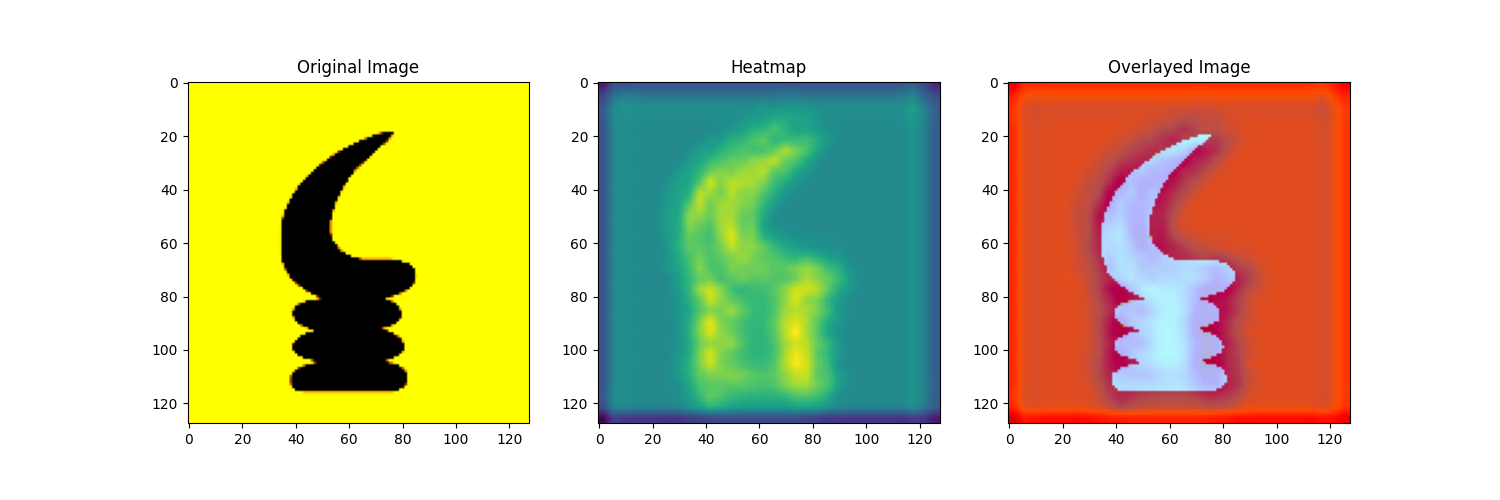}
    \end{subfigure}
    \vspace{-1.7em}
    
    \begin{subfigure}
        \centering
        \includegraphics[width=15cm]{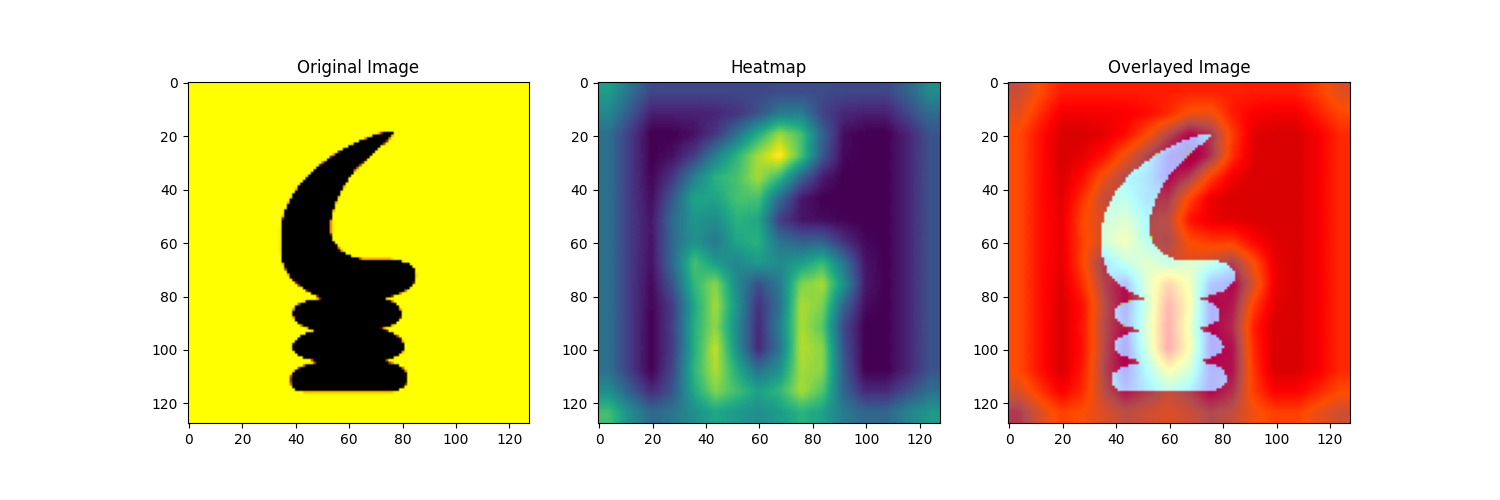}
    \end{subfigure}
    \caption{Visualizing the areas that significantly influence the model's predictions for interoperability: convo layer (\textbf{top}), activation layer (\textbf{Middle}), and pooling layer (\textbf{Bottom}).}
    \label{fig:adin_heatmap}
\end{figure}

For model interpretability, we resize, normalize and convert images to tensors, making them compatible with the CNN model, where a custom hook function is introduced to capture activations from the chosen target layers during the forward pass. We aggregate feature maps and apply the ReLU activation function to generate a heatmap that emphasizes essential regions. The resulting heatmap values are normalized to 0 and 1.
Finally, we overlay the heatmap on the original image with a specified weight to directly visualize the areas that significantly influence the model's predictions. Figure \ref{fig:adin_heatmap} efficiently displays the original image, heatmap, and overlaid image of the last layers before the fully connecting layer. Notably, we observe distinct feature characteristics in the last convo, activation, and pooling layers, plotted separately.


\section{Conclusion}\label{sec:conc}
In this research, we dived into classical machine learning and the power of deep learning models to tackle the complex task of classifying and recognizing Adinkra symbols. The idea led to a newly constructed ADINKRA dataset comprising 174,338 images organized into 62 distinct classes, each representing a singular and emblematic symbol. We used a CNN model for classification and recognition using six convolutional layers, three fully connected layers, and optional dropout regularization. We assess the model's performance by measuring its accuracy and convergence and visualizing the areas that significantly influence its predictions. The deep learning models generalize well by applying their learned knowledge to new, unseen data, reflecting the quality of the Adinkra data. The high-quality, diverse, and representative training data significantly impact the model's performance. As these evaluations establish a fundamental baseline for upcoming assessments of the ADINKRA dataset, we aim to offer an exhilarating challenge to machine learning experts interested in utilizing artificial intelligence's capabilities for comprehending and elucidating cultural heritage.



\bibliographystyle{unsrt} 

\newpage

\textbf{Supplementary Material}\label{sec:suppl} \\

\textbf{Adinkra Dataset}

The ADINKRA dataset is a remarkable collection that encapsulates the vibrant and diverse symbols of Ghanaian culture, painstakingly extracted from the vast canvas of the Google website. Each symbol within this dataset is a captivating 
\begin{figure}[!htbp]
     \centering
         \includegraphics[width=15cm]{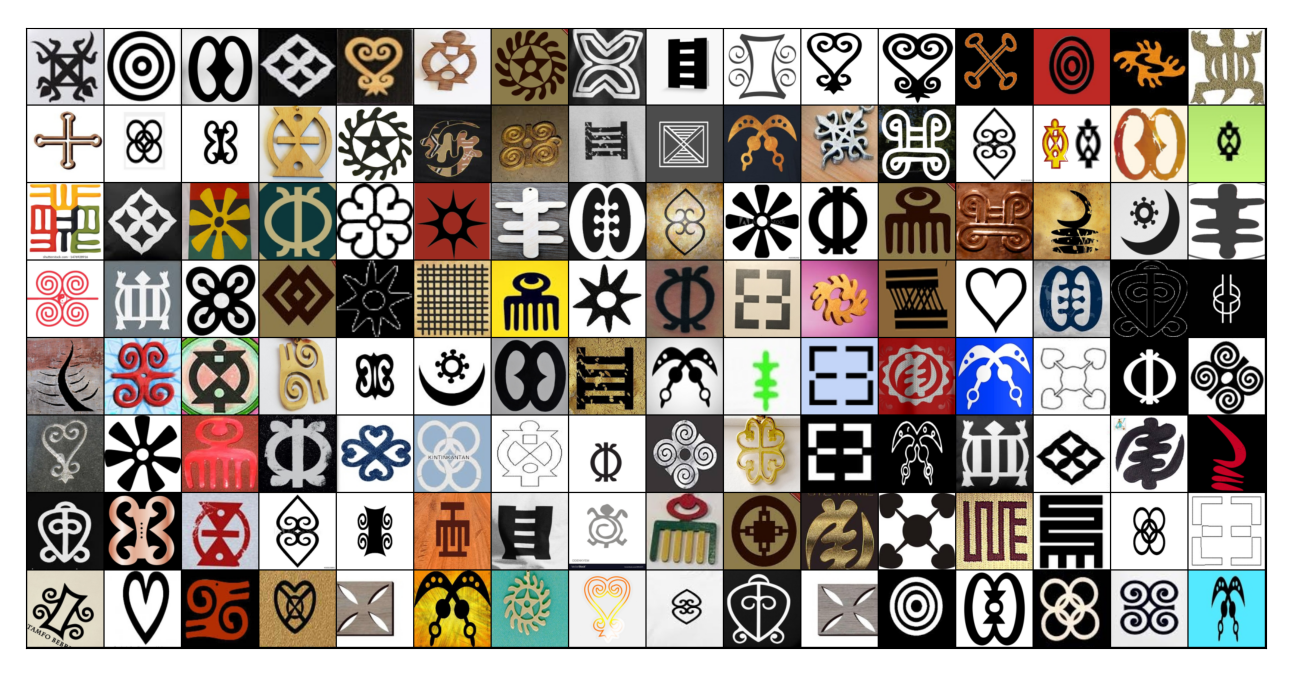}
         \label{fig:y equals x}
        \caption{Batch sample of the dataset representing the emblematic Adinkra symbols in different visual contexts}
    \label{fig:adin_sample}
\end{figure}
testament to the Ghanaian people's rich history, profound values, and enduring beliefs. These symbols (Figure \ref{fig:adin_sample}), each unique in its own right, serve as visual narratives that encapsulate the essence of a culture deeply rooted in tradition and symbolism. The dataset's original images are offered in grayscale and RGB color formats, comprehensively representing the symbols in different visual contexts. Including both formats enhances the versatility of the dataset, enabling researchers and practitioners to explore the intricacies of the symbols through various lenses.
The ADINKRA dataset comprises a substantial collection, with 139,471 images earmarked for training purposes. These images are meticulously curated and labeled, facilitating the development and evaluation of machine learning algorithms and models. Furthermore, the data set offers a robust evaluation set consisting of 34,867 test images, ensuring rigorous testing and validation of classification algorithms. Downloadable file names, descriptions, and the number of samples in the dataset are shown in Table \ref{adinkra:dataset}.

One of the distinguishing features of the ADINKRA dataset is its impressive diversity in the form of 62 distinct classes, each representing a unique symbol. These classes span a broad spectrum of cultural motifs, each with distinct significance and symbolism. The labels/classes of the dataset are names in the local dialect (Twi) and their meanings in English. This rich diversity makes the dataset a valuable resource for cultural studies. It presents an exciting challenge for machine learning practitioners who seek to harness the power of artificial intelligence to understand and interpret cultural heritage.
\begin{table}[!htbp] 
\caption{Descriptions and the number of samples of the ADINKRA dataset}
\label{adinkra:dataset} 
  \centering
  \label{tb1}
  \begin{tabular}{lccc}
\hline 
     \textbf{Names} & \textbf{Description} & \textbf{\# Samples} & \textbf{Size (bytes)} \\
    \hline
    \hline
    adinkra\_train\_sample.tar.xz & Train set images & 139,471 & 1,893,403,004 \\
    adinkra\_test\_sample.tar.xz & Test set images & 34,867  & 491,809,512 \\
    adinkra\_labels.txt & Classes  & 62  & 92 \\
\hline
  \end{tabular}
\end{table}

\end{document}